\documentclass{article} 
\usepackage{collas2026_conference,times}
\usepackage{easyReview}

\usepackage{amsmath,amsfonts,bm}

\def\eqref#1{equation~\ref{#1}}

\def\1{\bm{1}}

\DeclareMathAlphabet{\mathsfit}{\encodingdefault}{\sfdefault}{m}{sl}
\SetMathAlphabet{\mathsfit}{bold}{\encodingdefault}{\sfdefault}{bx}{n}

\usepackage{hyperref}
\usepackage{standalone}
\usepackage{multirow}
\usepackage{tabularx}
\usepackage{graphicx}
\usepackage{cleveref}

\usepackage{amsmath}
\usepackage{booktabs}
\usetikzlibrary{
  arrows.meta,
  calc,
  positioning,
  fit,
  backgrounds
}

\definecolor{stateblue}{HTML}{2F6BBA}
\definecolor{updateorange}{HTML}{D88922}
\definecolor{targetpurple}{HTML}{B83280}
\definecolor{darkink}{HTML}{222222}
\definecolor{softgray}{HTML}{F3F4F6}
\definecolor{linegray}{HTML}{8A8F98}
\definecolor{textgray}{HTML}{555B63}

\usepackage{hyperref}
\hypersetup{
    colorlinks=true,
    linkcolor=red,
    filecolor=magenta,
    urlcolor=blue,
    citecolor=purple,
    pdftitle={Overleaf Example},
    pdfpagemode=FullScreen,
    }

\title{Where Decoder Cosine Similarity Fails for SAE Feature Flow Discovery}

\author{Hendrik Droste$^{1}$, Christian Medeiros Adriano$^{1}$, Kathrin Korte$^{2}$, Holger Giese$^{1}$ \\
$^1$Hasso Plattner Institute, Germany, hendrik.droste@uni-potsdam.de\\
 $^2$IT University of Copenhagen, Denmark, kort@itu.dk
}

\preprintcopy 

\begin{document}

\maketitle

\begin{abstract}
Foundation models are increasingly adapted through fine-tuning, model editing, and alignment procedures while retaining previously acquired capabilities.
Understanding the internal computations that support these adaptations is therefore becoming increasingly important for continual model evolution.
Sparse autoencoders (SAEs) provide interpretable feature dictionaries for residual-stream activations and sublayer outputs, but it remains unclear how state features and update features interact to produce downstream residual features.
In this work, we focus on MLP updates as a first test case.
We construct a transition atlas of triples $s_k + u_j \rightarrow t_\ell$, where a residual-state feature and an MLP-update feature jointly predict a target residual feature, and validate candidate triples by ablating the decoded update feature.
In a 20M-token Pythia-160M \(L_7 \rightarrow L_8\) run, we find 38,125 strong ablation-effect transitions, but 88.0\% have both state-target and update-target decoder cosine similarity below 0.7.
As a preliminary cross-model check, a run of 20M-token Gemma-3-4B \(L_{21} \rightarrow L_{22}\) causally validates only the top 30,000 ranked candidate triples by ablating the decoded update feature, and 53.6\% of strong-effect triples have both state-target and update-target decoder cosine similarity below 0.7.
The Gemma result is directionally consistent with Pythia, but weaker, since update-target cosine recovers many of the strongest Gemma effects and the run is not a full-atlas causal validation.
Ultimately, our results suggest that feature flow atlases can serve as diagnostics of representation‑update mechanisms and thereby inform tools for steering model updates. Future work will validate more complex patterns across layers, models, and SAE families.

\end{abstract}

\section{Introduction}
Large foundation models are increasingly adapted after pretraining through fine-tuning, parameter-efficient methods, model editing, and incremental knowledge integration, rather than deployed as fixed systems \citep{zheng2025towards,zhao2026survey}.
These settings require incorporating new information while preserving prior capabilities, echoing the stability--plasticity trade-off in continual learning \citep{kirkpatrick2017overcoming,wickramasinghe2023continual}. 
Existing work can localize factual associations in feed-forward modules \citep{meng2023locatingeditingfactualassociations} and identify concepts that shift or emerge during chat tuning \citep{minder2025overcomingsparsityartifactscrosscoders}, but it offers less insight into the causal feature-level transformations by which internal computations are reused, refined, or suppressed.

Sparse autoencoders provide interpretable decompositions of transformer representations into sparse semantic features \citep{cunningham2023sparseautoencodershighlyinterpretable}, and recent work studies how such features persist, transform, or emerge across layers \citep{balcells2024evolutionsaefeatureslayers,laptev2025analyzefeatureflowenhance}.
We therefore treat feature flow discovery as a mechanistic step toward understanding how residual-state features and MLP-update features interact during computation.

We construct a transition atlas of triples \(s_k + u_j \rightarrow t_\ell\), where a residual-state feature \(s_k\) and an MLP-update feature \(u_j\) jointly predict a downstream residual feature \(t_\ell\).
With this atlas we answer the following \textbf{research question}:
Are strong cross-layer feature transitions recoverable from pairwise decoder geometry alone, or do causal transitions depend on the joint presence of a prior residual-state feature and an MLP-update feature?

\section{Feature Flow Discovery}

Our feature flow atlas contains transitions in the form of \(s_k + u_j \rightarrow t_\ell\), where $s_k$ is the source residual state feature, $u_j$ is the target-layer MLP update feature, and $t_\ell$ is the target residual feature.
We use decoder cosine similarity as a pairwise geometry baseline for the atlas.
For each atlas triple, we compute the state-target decoder cosine \(c_{s,t}\) and the update-target decoder cosine \(c_{u,t}\).
Following \cite{laptev2025analyzefeatureflowenhance}, we identify the most cosine-similar state and update features for each target feature.
Together, these comparisons ask whether the pairwise decoder cosine alone would recover the same transitions found by the atlas.

\subsection{Matching Pipeline}

\begin{figure}[ht]
    \centering
    \begin{tikzpicture}[
  font=\sffamily,
  >=Latex,
  panel/.style={
    draw=linegray,
    rounded corners=3pt,
    line width=0.5pt,
    fill=white
  },
  box/.style={
    draw=linegray,
    rounded corners=2pt,
    line width=0.45pt,
    fill=softgray,
    align=center,
    inner sep=4pt,
    font=\sffamily\small
  },
  smallbox/.style={
    draw=linegray,
    rounded corners=2pt,
    line width=0.4pt,
    fill=white,
    align=center,
    inner sep=3pt,
    font=\sffamily\scriptsize
  },
  feature/.style={
    draw,
    rounded corners=2pt,
    line width=0.8pt,
    fill=white,
    align=center,
    inner sep=4pt,
    font=\sffamily\small\bfseries
  },
  note/.style={
    align=center,
    text=textgray,
    font=\sffamily\scriptsize
  },
  arrow/.style={
    -{Latex[length=2.2mm]},
    line width=0.65pt,
    draw=darkink
  },
  faintarrow/.style={
    -{Latex[length=1.8mm]},
    line width=0.45pt,
    draw=linegray
  },
  feedline/.style={
    line width=0.55pt,
    draw=linegray
  }
]

\node[panel, minimum width=7.35cm, minimum height=7.55cm] (panelA) at (0,-0.25) {};

\coordinate (streamTop) at (-2.55,2.35);
\coordinate (stateHook) at (-2.55,1.86);
\coordinate (attnInputTap) at (-2.55,1.62);
\coordinate (resIn) at (-2.55,1.38);
\coordinate (addAttn) at (-2.55,0.15);
\coordinate (mlpInputTap) at (-2.55,-0.32);
\coordinate (addMlp) at (-2.55,-1.75);
\coordinate (resOut) at (-2.55,-2.85);
\coordinate (streamBottom) at (-2.55,-3.6);

\draw[faintarrow, dashed] (streamTop) -- node[left, note, xshift=-2pt] {earlier\\layers} (stateHook);
\draw[feedline] (stateHook) -- (resIn);
\draw[feedline] (resIn) -- node[left, note, xshift=-2pt] {$r_{i-1}$} (addAttn);
\draw[feedline] (addAttn) -- (addMlp);
\draw[feedline] (addMlp) -- node[left, note, xshift=-2pt] {$r_i$} (resOut);
\draw[faintarrow, dashed] (resOut) -- node[left, note, xshift=-2pt] {later\\layers} (streamBottom);

\node[circle, draw=linegray, fill=white, inner sep=1.4pt, font=\scriptsize] (plusAttn) at (addAttn) {$+$};
\node[circle, draw=linegray, fill=white, inner sep=1.4pt, font=\scriptsize] (plusMlp) at (addMlp) {$+$};

\node[smallbox, minimum width=1.65cm, minimum height=0.62cm] (attnBlock) at (-0.75,1.08)
  {Attention$_i$};
\node[smallbox, minimum width=1.65cm, minimum height=0.62cm] (mlpBlock) at (-0.75,-0.95)
  {MLP$_i$};
\coordinate (mlpOutput) at (mlpBlock.south);
\coordinate (mlpSaeTap) at ($(mlpOutput)+(0,-0.32)$);

\draw[faintarrow] (attnInputTap) -| (attnBlock.north);
\draw[faintarrow] (mlpInputTap) -| (mlpBlock.north);
\draw[feedline] (attnBlock.south) |- node[pos=0.22, left, note] {$a_i$} (plusAttn.east);
\draw[feedline] (mlpOutput) |- node[pos=0.22, left, note] {$m_i$} (plusMlp.east);

\fill[stateblue] (stateHook) circle (2.2pt);
\fill[updateorange] (mlpSaeTap) circle (1.7pt);
\fill[targetpurple] (resOut) circle (2.2pt);

\node[feature, draw=stateblue, text=stateblue, minimum width=2.2cm]
  (stateFeat) at (2.05,1.86)
  {RES SAE$_{i-1}$\\[-1pt]$\rightarrow\ z_{s_k}$};
\node[feature, draw=updateorange, text=updateorange, minimum width=2.2cm]
  (updateFeat) at ($(mlpSaeTap)+(2.8,0)$)
  {MLP SAE$_i$\\[-1pt]$\rightarrow\ z_{u_j}$};
\node[feature, draw=targetpurple, text=targetpurple, minimum width=2.2cm]
  (targetFeat) at (2.05,-2.85)
  {RES SAE$_i$\\[-1pt]$\rightarrow\ z_{t_\ell}$};

\draw[arrow, draw=stateblue] (stateHook) -- node[above, note, text=stateblue] {encode} (stateFeat.west);
\draw[arrow, draw=updateorange] (mlpSaeTap) -- node[above, note, text=updateorange] {encode} (updateFeat.west);
\draw[arrow, draw=targetpurple] (resOut) -- node[above, note,text=targetpurple] {encode} (targetFeat.west);

\node[panel, minimum width=7.0cm, minimum height=7.55cm] (panelB) at (8.0,-0.25) {};

\node[feature, draw=stateblue, text=stateblue] (bState) at (6.05,2.65) {$z_{s_k}$};
\node[feature, draw=updateorange, text=updateorange] (bUpdate) at (8.0,2.65) {$z_{u_j}$};
\node[feature, draw=targetpurple, text=targetpurple] (bTarget) at (9.95,2.65) {$\Delta z_{t_\ell}$};

\node[box, fill=white, text width=5.2cm, minimum height=0.9cm]
  (triple) at (8.0,1.48)
  {\textbf{candidate triple}\\[-1pt]
   $s_k + u_j \rightarrow \Delta z_{t_\ell}$};

\draw[arrow, draw=stateblue] (bState) -- (triple);
\draw[arrow, draw=updateorange] (bUpdate) -- (triple);
\draw[arrow, draw=targetpurple] (bTarget) -- (triple);

\node[box, fill=softgray, text width=5.75cm, minimum height=2.05cm]
  (score) at (8.0,-0.55)
  {\textbf{score triple}\\[-1pt]
   \scriptsize
   \begin{tabular}{@{}r@{\ }c@{\ }l@{}}
     joint & $=$ & $\mathbb{E}[z^{(i)}_\ell\mid s_k,u_j]$\\
     state & $=$ & $\mathbb{E}[z^{(i)}_\ell\mid s_k]$\\
     update & $=$ & $\mathbb{E}[z^{(i)}_\ell\mid u_j]$\\
     global & $=$ & $\mathbb{E}[z^{(i)}_\ell]$\\[2pt]
     \multicolumn{3}{@{}c@{}}{
       $\mathrm{effect}=\text{joint}-\text{state}-\text{update}+\text{global}$
     }\\[1pt]
     \multicolumn{3}{@{}c@{}}{
       $\mathrm{score}=\mathrm{effect}/(\operatorname{std}(z^{(i)}_\ell)+\epsilon)$
     }
   \end{tabular}};

\node[smallbox, fill=white, text width=5.25cm, minimum height=1.32cm]
  (table) at (8.0,-3.0)
  {\textbf{ranked atlas}\\[2pt]
   \begin{tabular}{@{}cccc@{}}
     \textbf{state} & \textbf{update} & \textbf{target} & \textbf{score}\\
     $s_{42}$ & $u_{8}$ & $t_{91}$ & 3.2\\
     $s_{17}$ & $u_{31}$ & $t_{12}$ & 2.7
   \end{tabular}};

\draw[faintarrow] (triple) -- (score);
\draw[faintarrow] (score) -- (table);

\node[panel, minimum width=7.0cm, minimum height=7.55cm] (panelC) at (15.6,-0.25) {};

\node[box, fill=white, minimum width=5.1cm, minimum height=0.75cm]
  (selected) at (15.6,2.55) {selected triple: $s_k + u_j \rightarrow t_\ell$};

\node[smallbox, text width=3.55cm, minimum height=0.75cm] (active) at (15.6,1.35)
  {tokens where\\$s_k$ and $u_j$ are active};

\node[box, fill=softgray, text width=5.15cm, minimum height=1.0cm]
  (ablate) at (15.6,0.05)
  {remove decoded update feature\\
   $\widetilde{m}_{i}=m_{i}-z_{u_j} d_{u_j}$};

\node[box, fill=white, text width=5.15cm, minimum height=0.95cm]
  (reencode) at (15.6,-1.35)
  {patch residual stream and re-encode\\with residual SAE$_i$};

\node[feature, draw=targetpurple, text=targetpurple, minimum width=5.1cm]
  (outcome) at (15.6,-2.85)
  {does target $t_\ell$ change\\by at least two standard deviations?};

\draw[faintarrow] (selected) -- (active);
\draw[arrow] (active) -- (ablate);
\draw[faintarrow] (ablate) -- (reencode);
\draw[arrow, draw=targetpurple] (reencode) -- (outcome);

\draw[arrow] ($(panelA.east)+(0.1,0)$) -- ($(panelB.west)+(-0.1,0)$);
\draw[arrow] ($(panelB.east)+(0.1,0)$) -- ($(panelC.west)+(-0.1,0)$);

\end{tikzpicture}
\caption{
Overview of the feature matching pipeline.
SAEs trained on the residual stream before layer \(i\), the MLP output of layer \(i\), and the residual stream after layer \(i\) define three separate feature spaces: state features \(s_k\), update features \(u_j\), and target features \(t_\ell\), where \(k\), \(j\), and \(\ell\) index features within their respective SAE dictionaries. 
Activation frequencies over tokens are used to score triples \(s_k + u_j \rightarrow t_\ell\), measuring whether update feature \(u_j\) predicts a target-feature change beyond what is expected from state feature \(s_k\) alone. 
High-scoring triples are then validated by ablating the decoded update feature and checking whether the target feature changes by at least two target-feature standard deviations.
}
    \label{fig:matching-pipeline}
\end{figure}
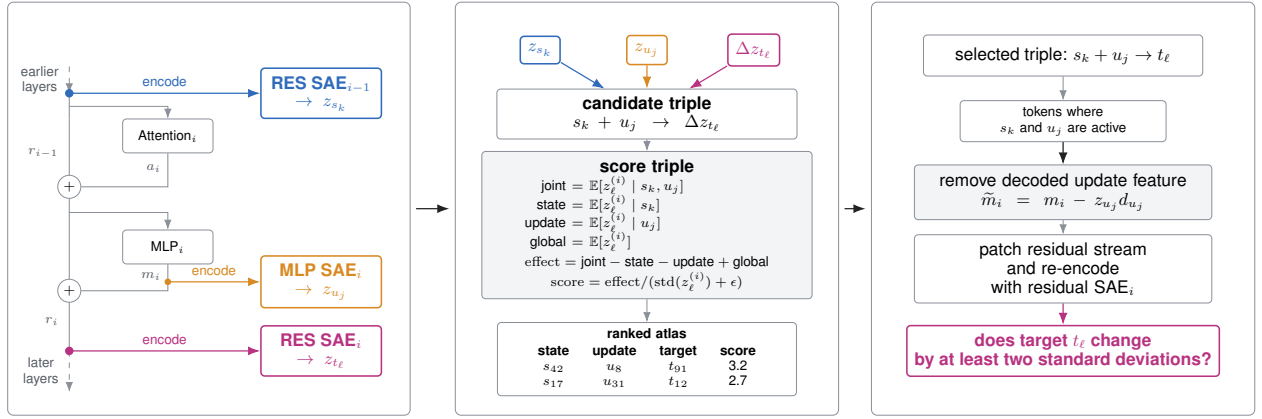

We created the atlas using a three-stage feature matching pipeline.
It requires SAEs trained for the state, update and target features.
In our case we trained TopK SAEs on the residual stream after layers 7 and 8 as well as the MLP output of layer 8 of the Pythia-160M model \citep{biderman2023pythiasuiteanalyzinglarge}.
To train stable SAEs, we adapted the training configuration from \cite{song2025positionmechanisticinterpretabilityprioritize}.
The hyperparameters are listed in Appendix~\Cref{app:sae-training-hyperparameters}.

Stage one of the pipeline is sparse event extraction.
We run the model and collect the source residual stream,  MLP-sublayer, target residual stream activations, and encode these with the corresponding SAEs.
To reduce runtime we only store the top 64 active features per token.

The second stage constructs a feature-update candidate atlas by reading the sparse activations and turning them into ranked transition triples.
We estimate, for each layer-\(i\) residual SAE feature, how much the layer-\(i\) MLP changes that feature.
To do this, we encode both the final residual stream and the pre-MLP residual stream in the same layer-\(i\) residual SAE basis:
\[
z^{(i)}_{\mathrm{after}}
=
E^{(i)}_{\mathrm{res}}\!\left(r^{(i)}_{\mathrm{post}}\right),
\qquad
z^{(i)}_{\mathrm{no\text{-}mlp}}
=
E^{(i)}_{\mathrm{res}}\!\left(r^{(i)}_{\mathrm{post}} - m^{(i)}\right),
\]
and define the MLP-induced target-feature change as
\[
\Delta z^{(i,\mathrm{mlp})}
=
z^{(i)}_{\mathrm{after}}
-
z^{(i)}_{\mathrm{no\text{-}mlp}} .
\]
If the MLP sublayer does not change target feature \(t_\ell\), then \(\Delta z^{(i,\mathrm{mlp})}_\ell = 0\).

Next, we accumulate pairwise baseline frequencies for state-target and update-target relationships.
These baselines prevent the atlas from ranking triples only because their individual features are common.
Only after measuring them can we ask whether the full triple \((s_k, u_j, t_\ell)\) exists in which all the features could be causally related.

After accumulating the pairwise baselines and joint triple frequencies, we convert each candidate triple into a normalized score.
For this, we compute 
\[
\mathrm{interaction\_score}(s_k, u_j, t_\ell)
=
\frac{
\mathbb{E}[z^{(i)}_\ell \mid s_k,u_j]
-
\mathbb{E}[z^{(i)}_\ell \mid s_k]
-
\mathbb{E}[z^{(i)}_\ell \mid u_j]
+
\mathbb{E}[z^{(i)}_\ell]
}{
\operatorname{std}(z^{(i)}_\ell) + \epsilon
},
\]

Finally, we filter low-support triples using minimum activation and co-occurrence thresholds, apply the pruning procedure described in Appendix~\Cref{app:pruning}, annotate triples with residual-feature component labels, and write the ranked atlas to disk.

Stage three validates the triples using ablation.
For each selected triple \((s_k, u_j, t_\ell)\), we rerun the model and identify token positions where both the state feature \(s_k\) and the MLP update feature \(u_j\) are active.
We then ablate only the decoded contribution of \(u_j\) from the MLP output, reconstruct the patched residual stream and re-encode it with the layer-\(i\) residual SAE.
The normalized ablation score measures the ablation effect in units of the target feature's own standard deviation:
\[
A_z(s_k, u_j, t_\ell)=
\frac{
\mathbb{E}[z^{(i)}_\ell \mid s_k,u_j]
-
\mathbb{E}[\widetilde{z}^{(i)}_\ell \mid s_k,u_j]
}{
\operatorname{std}(z^{(i)}_\ell)+\epsilon
}.
\]
As a smaller complementary experiment, we ablate selected state features \(s_k\) for high-ranked \((s_k,u_j)\) pairs and measure the normalized effect on the corresponding update feature \(u_j\).

We normalize ablation effects because SAE feature activations are not naturally comparable across features.
Some feature activations are much larger or more variable than others \citep{sun2024massiveactivationslargelanguage}.
Without normalization, high-scale target features would dominate the validation results even when their relative causal effect is small.

\subsection{Cosine Matching}
Decoder cosine similarity provides a natural geometric baseline because an SAE decoder vector represents the direction in activation space that a feature contributes.
A high cosine similarity, therefore, indicates that two features write in similar directions, independent of their vector magnitudes.
Moreover, cosine similarity can be computed directly from the SAE weights without collecting activation statistics or performing interventions.
The issue with cosine matching, described by \cite{balcells2024evolutionsaefeatureslayers}, is that feature-similarity measures cannot establish that an upstream feature causally contributes to a downstream feature.
It measures whether two SAE features write in similar directions in activation space.
It does not test whether the features are active on the same tokens, whether one feature actually changes the other, or whether the relation depends on a previous residual stream state.
Thus, two individually low-cosine features can still jointly create a target feature when their combined contributions move the residual stream into the target feature's activation region.

\section{Preliminary Results}
The 20M-token Pythia-160M \(L_7 \rightarrow L_8\) run produced 262,735 filtered MLP transition triples.
Of these, 38,125 have a strong ablation effect under the criterion \(|A_z| \geq 2\), corresponding to 14.5\% of the filtered triples.
We use these validated triples to test whether strong feature transitions can be recovered from pairwise decoder geometry alone.

Our results show that high update-target cosine similarity captures direct MLP write relations.
For \(c_{u,t} \geq 0.7\), 2,254 of 3,211 triples reach \(|A_z| \geq 2\) (70.2\%).
This trend remains visible at lower cosine thresholds.
For \(c_{u,t} \geq 0.5\), 3,876 of 5,855 triples reach \(|A_z| \geq 2\) (66.2\%).
This confirms the findings by \cite{laptev2025analyzefeatureflowenhance}.
Based on the high cosine similarity, we hypothesize that in triples with high update-target cosine similarity, a new feature is created by the update feature.
The update-to-target results above measure whether ablating the update feature changes the target feature.
We also run a smaller-sized state-to-update experiment, where the state feature is ablated, and the MLP update feature is measured.
Across the 4,096 highest-ranked unique \((s,u)\) pairs and 500k tokens, 3,765 pairs with at least ten active tokens reach \(A_z \geq 2\) (91.9\%), with a median update-feature effect of \(A_z=40.6\).
This suggests that state features can strongly condition MLP update features, motivating the triple-level analysis below.

\begin{figure}[t]
    \centering
    \begin{minipage}[t]{0.49\textwidth}
        \centering
        \includegraphics[width=\linewidth]{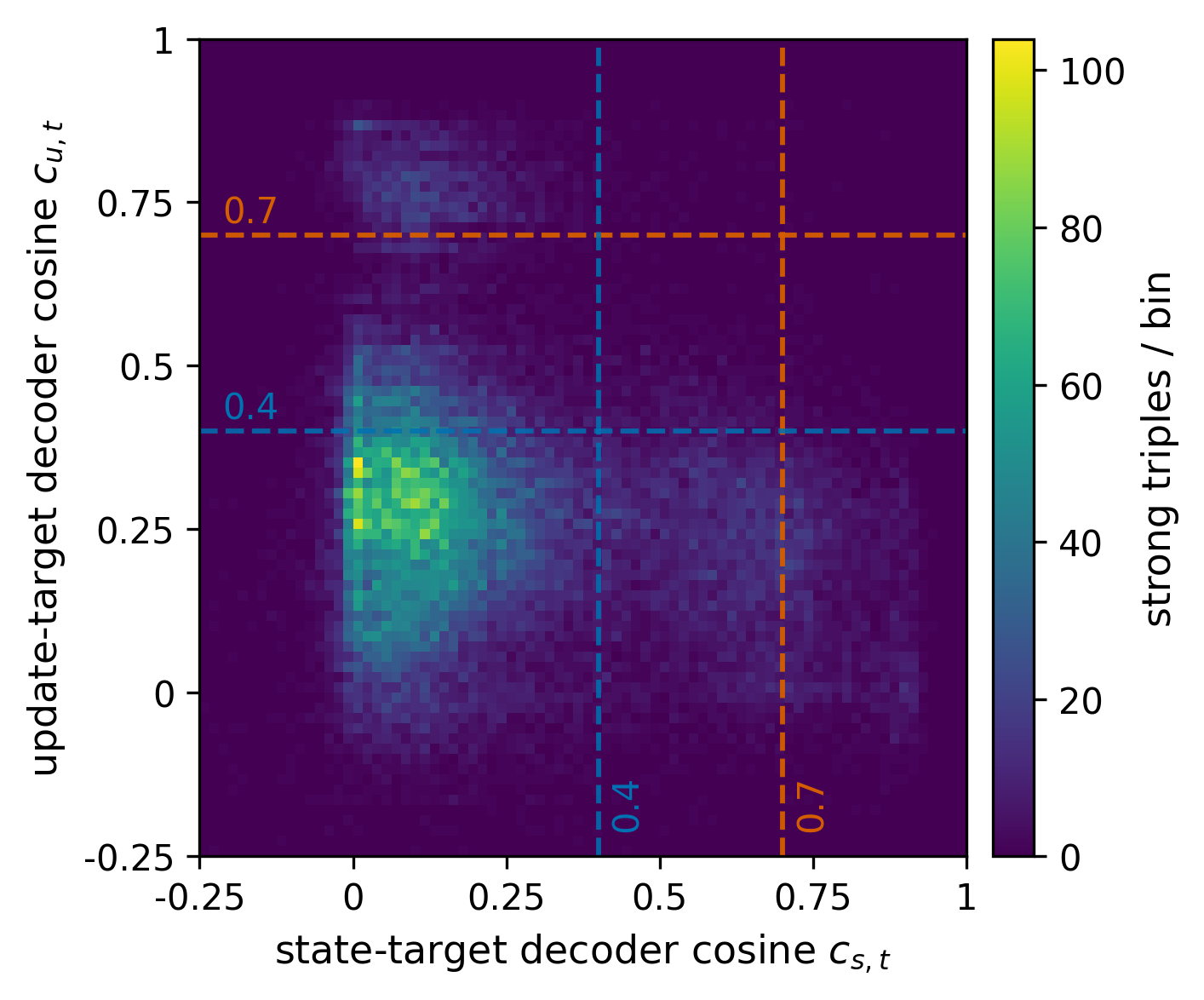}
        {\small\textbf{(a) Pythia-160M full atlas}\par}
    \end{minipage}\hfill
    \begin{minipage}[t]{0.49\textwidth}
        \centering
        \includegraphics[width=\linewidth]{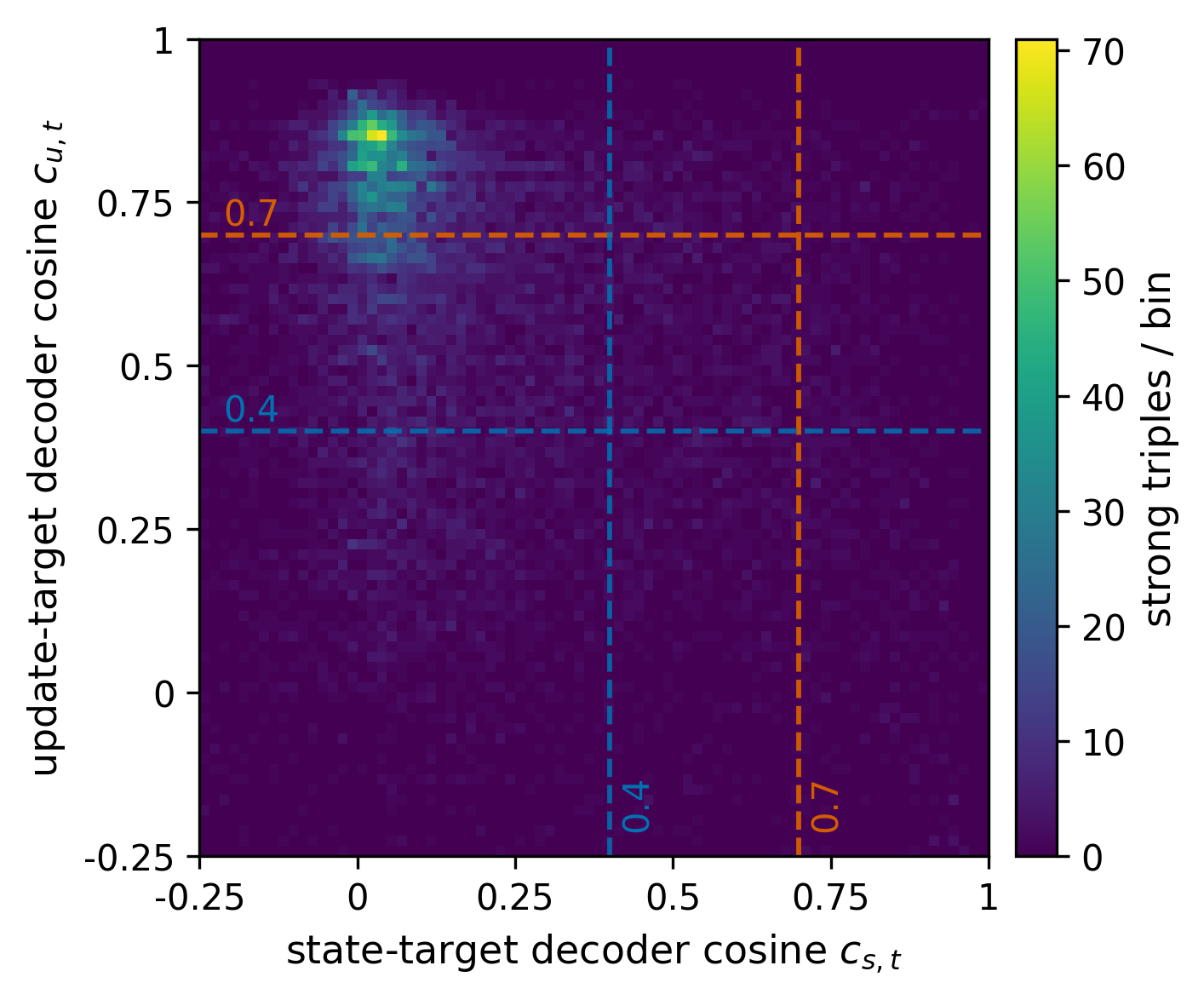}
        {\small\textbf{(b) Gemma-3-4B ranked subset}\par}
    \end{minipage}
    \caption{
    Strong causal (\(|A_z| \geq 2\)) transition triples in decoder-cosine space.
    Panel (a) shows the Pythia-160M \(L_7 \rightarrow L_8\) full-atlas run.
    Panel (b) shows the Gemma-3-4B \(L_{21} \rightarrow L_{22}\) validation subset, restricted to the top 30,000 ranked candidates.
    Dashed lines mark pairwise cosine thresholds \(\tau=0.4\) and \(\tau=0.7\).
    The Pythia panel shows that most strong transitions are missed by high pairwise-cosine thresholds; the Gemma panel provides a qualified cross-model check rather than a full-atlas recall measurement.
    }
    \label{fig:strong-causal-triples-cosine-space}
    \label{fig:gemma-strong-causal-triples-cosine-space}
\end{figure}

However, the pairwise decoder's cosine similarity misses the strongest causal triples. 
Of the 38,125 triples with strong ablation effects, 33,550 have both \(c_{s,t}<0.7\) and \(c_{u,t}<0.7\), meaning that 88.0\% of strong causal transitions would be missed by a high-cosine threshold on either the state-target or update-target pair. 
This remains true even when the threshold is substantially lowered.
At \(\tau=0.4\), 23,677 of 38,125 strong-effect triples still have both pairwise cosine scores below threshold (62.1\%).
We use \(\tau=0.4\) and \(\tau=0.7\) as reference thresholds following \cite{paulo2025sparseautoencoderstraineddata}.
Figure~\ref{fig:gemma-strong-causal-triples-cosine-space} shows that the validated triples are not concentrated in the corresponding high-cosine regions.
Instead, the dominant mass lies at a low state-target cosine, with \(c_{s,t}\) close to zero, and only a moderate update-target cosine. 
Thus, for many validated transitions, neither the previous residual feature nor the MLP update feature is individually a high-cosine match to the target feature. 
The plot also shows a smaller high-\(c_{u,t}\) band, which is consistent with direct write-like transitions where the update feature decoder is already aligned with the target decoder. 
However, this direct-write regime is only one subpopulation. 
Most strong causal transitions occupy lower-cosine regions, indicating that their relationship is better described as a state-conditioned interaction than as a pairwise geometric match. 
This supports the need for a transition atlas that scores triples \(s_k + u_j \rightarrow t_\ell\), rather than relying only on pairwise decoder cosine similarity.
Thus, in the full Pythia atlas, the answer to our research question is negative.
Pairwise decoder cosine alone does not recover most strong causal feature transitions.

As a cross-model check, we evaluated Gemma-3-4B \(L_{21} \rightarrow L_{22}\) using pretrained Gemma Scope residual and MLP SAEs \citep{mcdougall2025gemma,gemmateam2025gemma3technicalreport}.
The atlas contains 231,198 filtered MLP triples, but to reduce runtime, we only validated the top 30,000 causal triples, so this is not a full-atlas validation.
Restricting the analysis to strong-effect triples, there are 12,139 candidates with \(|A_z| \geq 2\).
Among these strong causal candidates, 6,506 (53.6\%) have both \(c_{s,t}<0.7\) and \(c_{u,t}<0.7\), and 2,291 (18.9\%) still have both pairwise cosine scores below 0.4.
A top-30k update-target cosine baseline overlaps with 8,278 strong-effect candidates (68.2\%) and misses 3,861 (31.8\%), while the state-target baseline overlaps with only 1,676 (13.8\%).
Thus, the Gemma result is weaker than the Pythia result.
It still shows that many strong triples are not recovered by state-target cosine, but update-target cosine explains a much larger fraction of the strongest effects, and the subset validation prevents a full-atlas recall claim.

\section{Limitations and Ongoing Work}
Our strongest evidence is the completed \(L_7 \rightarrow L_8\) Pythia-160M full-atlas run using custom-trained SAEs.
The Gemma-3-4B result provides only qualified cross-model support, since it causally validates only the top 30,000 ranked triples, and its strongest effects are more often recovered by update-target cosine.
It therefore remains open how much the observed low recall of decoder cosine similarity depends on model scale, layer choice, SAE architecture, pruning parameters, and validation budget.

We use these limitations to set the priorities for our future work. As we currently only have estimates of how different pruning parameters affect the atlas, we plan to compare additional scoring functions for ranking triples before causal validation. We also aim to validate more Gemma Scope~\citep{gemmateam2025gemma3technicalreport} layer transitions to test whether context‑conditioned feature‑flow relations appear consistently across layers. Besides improving scalability, the usage of pretrained SAEs could also enable the adoption of tools like Neuronpedia \citep{neuronpedia} to gain deeper insight into SAE features.

Creating a feature‑flow atlas for one or more LLMs would open up many opportunities to generate new hypotheses about how individual features work. Although this study focuses on inference dynamics within a fixed model, we believe feature‑flow analysis could also become a useful tool for studying learning dynamics in continually adapted foundation models. Existing continual‑learning approaches mainly analyze parameter updates or representation‑level changes, whereas feature‑flow gives access to the computational interactions that drive representation evolution. Comparing feature‑flow atlases across successive stages of adaptation may therefore offer a complementary mechanistic perspective on transfer, interference, and knowledge integration in foundation models.

\bibliography{references}
\bibliographystyle{collas2026_conference}

\appendix
\section{Appendix}

\subsection{Notation and Score Definitions}
\label{app:notation-score-definitions}

\subsubsection{Symbols}

\begin{tabularx}{\textwidth}{lX}
\toprule
Symbol & Meaning \\
\midrule
\(i\) & Target transformer layer. \\
\(r^{(i-1)}_{\mathrm{post}}\) & Residual stream before the target layer. \\
\(r^{(i)}_{\mathrm{post}}\) & Residual stream after the target layer. \\
\(a^{(i)}\) & Attention output of layer \(i\). \\
\(m^{(i)}\) & MLP output of layer \(i\). \\
\(\widetilde{m}^{(i)}\) & MLP output after removing the decoded update feature. \\
\(E^{(i)}_{\mathrm{res}}\) & Residual-stream SAE encoder for layer \(i\). \\
\(E^{(i)}_{\mathrm{mlp}}\) & MLP-output SAE encoder for layer \(i\). \\
\(s_k\) & Layer-\((i-1)\) residual state feature. \\
\(u_j\) & Layer-\(i\) MLP update feature. \\
\(t_\ell\) & Layer-\(i\) residual target feature. \\
\(z^{(i)}_\ell\) & Activation of target feature \(t_\ell\) after layer \(i\). \\
\(\widetilde{z}^{(i)}_\ell\) & Target-feature activation after patching. \\
\(\Delta z^{(i,\mathrm{mlp})}_\ell\) & MLP-induced change in target feature \(t_\ell\). \\
\(d_{s_k}, d_{u_j}, d_{t_\ell}\) & Decoder directions for the state, update, and target features. \\
\(A_z\) & Normalized ablation effect. \\
\(c_{s,t}, c_{u,t}\) & State-target and update-target decoder-cosine baselines. \\
\(\epsilon\) & Small constant for numerical stability. \\
\(\tau\) & Pairwise decoder-cosine threshold, e.g. 0.4 or 0.7. \\
\bottomrule
\end{tabularx}

\subsubsection{Layer Decomposition}

For each token, the residual stream after layer \(i\) is approximated by
\[
r^{(i)}_{\mathrm{post}}
\approx
r^{(i-1)}_{\mathrm{post}} + a^{(i)} + m^{(i)}.
\]

\subsubsection{MLP-Induced Target Change}

We encode both the original residual stream and the no-MLP residual stream in the same layer-\(i\) residual SAE basis:
\[
z^{(i)}_{\mathrm{after}}
=
E^{(i)}_{\mathrm{res}}\!\left(r^{(i)}_{\mathrm{post}}\right),
\qquad
z^{(i)}_{\mathrm{no\text{-}mlp}}
=
E^{(i)}_{\mathrm{res}}\!\left(r^{(i)}_{\mathrm{post}} - m^{(i)}\right).
\]
The MLP-induced target-feature change is
\[
\Delta z^{(i,\mathrm{mlp})}
=
z^{(i)}_{\mathrm{after}}
-
z^{(i)}_{\mathrm{no\text{-}mlp}}.
\]

\subsubsection{Decoder-Cosine Baselines} 

Let \(d_{s_k}\), \(d_{u_j}\), and \(d_{t_\ell}\) be decoder directions for the state, update, and target features. The cosine baselines are
\[
c_{s,t}
=
\frac{d_{s_k}^{\top} d_{t_\ell}}
{\|d_{s_k}\|_2 \|d_{t_\ell}\|_2},
\qquad
c_{u,t}
=
\frac{d_{u_j}^{\top} d_{t_\ell}}
{\|d_{u_j}\|_2 \|d_{t_\ell}\|_2}.
\]

\subsubsection{Causal Validation}

For a selected triple \((s_k,u_j,t_\ell)\), we validate the candidate transition token positions where both \(s_k\) and \(u_j\) are active.
We remove the decoded update-feature contribution from the MLP output,
\[
z^{(i,\mathrm{mlp})}_{u_j} d^{(i,\mathrm{mlp})}_{u_j},
\]
and reconstruct the patched residual stream
\[
\widetilde{r}^{(i)}_{\mathrm{post}}
=
r^{(i-1)}_{\mathrm{post}} + a^{(i)} + \widetilde{m}^{(i)}.
\]
Let \(\widetilde{z}^{(i)}_\ell\) be the target-feature activation after patching.
The normalized update-to-target ablation effect is
\[
A_z(s_k, u_j, t_\ell)
=
\frac{
\mathbb{E}[z^{(i)}_\ell \mid s_k,u_j]
-
\mathbb{E}[\widetilde{z}^{(i)}_\ell \mid s_k,u_j]
}{
\operatorname{std}(z^{(i)}_\ell) + \epsilon
}.
\]
A triple is counted as \emph{validated} when its intervention produces a large normalized target-feature change, using the magnitude-only criterion \(|A_z| \geq 2\).
This is the validation criterion used throughout the main analysis, because the question there is whether pairwise decoder geometry recovers transitions with large intervention effects.
We do not require the atlas score to predict the sign of the ablation effect.

\subsubsection{State-to-Update Check}
As a complementary check, we also test whether state features causally condition MLP update features.
For selected \((s_k,u_j)\) pairs, we ablate the decoded state-feature contribution and measure the normalized change in the corresponding MLP SAE update feature \(u_j\).
This check validates the causal relation \(s_k \rightarrow u_j\), showing that the MLP update feature can depend on the prior residual-state feature.

\subsection{Pruning}
\label{app:pruning}
A full enumeration of all possible transition triples is computationally infeasible.
In principle, each token could activate many SAE features with different strengths, so the search space per token scales with the product of the layer-\((i-1)\) residual SAE dimension, the layer-\(i\) MLP-out SAE dimension, and the layer-\(i\) residual SAE dimension.
We therefore restrict the token-level search space using the number of layer-\((i-1)\) state features, layer-\(i\) update features, and layer-\(i\) target features retained per token as parameters.
For the Pythia run, we retain the top 64 active features per token during event extraction, and for Gemma we validate only the top 30,000 ranked causal triples.
For each token, we keep the top active state and update features, and a larger set of candidate target features.
The raw candidate set is then the Cartesian product of these sparse feature sets.
We further prune this set by retaining only the strongest \(n\) local candidate triples per token, ranked by
\[
\left|z_{s_k} \cdot z_{u_j} \cdot \Delta z^{(i,\mathrm{mlp})}_\ell\right|,
\]
where \(z_{s_k}\) is the layer-\((i-1)\) state activation, \(z_{u_j}\) is the layer-\(i\) MLP update activation, and \(\Delta z^{(i,\mathrm{mlp})}_\ell\) is the MLP-induced change in the layer-\(i\) target feature.

This token-level pruning is separate from the later validation budget, which determines how many ranked triples are interventionally tested.
With this pruning, the candidate atlas should be understood as a precision-oriented discovery atlas rather than an exhaustive enumeration of all possible feature transitions.

\begin{table}[ht]
\centering

\label{tab:pipeline-validation-summary}
\small
\begin{tabular}{lrr}
\toprule
\textbf{Quantity} & \textbf{Pythia-160M \(L_7 \rightarrow L_8\)} & \textbf{Gemma-3-4B \(L_{21} \rightarrow L_{22}\)} \\
\midrule
Atlas/scoring tokens & 20M & 20M \\
Unfiltered MLP candidate triples & 199,004,232 & 146,283,098 \\
Filtered MLP triples & 262,735 & 231,198 \\
Validation tokens & 4M & 4M \\
Validated interaction-ranked triples & 262,735 & 30,000 \\
Strong ablation-effect triples (\(|A_z| \geq 2\)) & 38,125 & 12,139 \\
\bottomrule
\end{tabular}
\caption{Pipeline and validation summary for the Pythia-160M full-atlas run and the Gemma-3-4B ranked validation subset.}
\end{table}

\subsection{Pythia SAE Training Hyperparameters}
\label{app:sae-training-hyperparameters}

The table lists the training configuration for the custom Pythia-160M SAEs.
For Gemma, we use pretrained width-16k, \(L_0\)-big Gemma Scope 2 residual SAEs at layers 21 and 22 and the layer-22 MLP SAE.

\begin{table}[ht]
\centering
\label{tab:sae-training-hyperparameters}
\begin{tabular}{ll}
\toprule
\textbf{Hyperparameter} & \textbf{Value} \\
\midrule
Model & \texttt{EleutherAI/pythia-160m} \\
Dataset & \texttt{monology/pile-uncopyrighted} \\
SAE architecture & TopK SAE \\
Dictionary size \(d_{\mathrm{SAE}}\) & 33024 \\
Input dimension \(d_{\mathrm{model}}\) & 768 \\
TopK sparsity \(K_{\mathrm{top}}\) & 100 \\
Training tokens & 8B \\
Context length & 128 \\
Token batch size & 4096 \\
Learning rate & \(3 \cdot 10^{-4}\) \\
Adam betas & \((0.9, 0.999)\) \\
Learning-rate warmup & 500 steps \\
Auxiliary loss coefficient & 1.0 \\
Training seed & 43 \\
\bottomrule
\end{tabular}
\caption{Training hyperparameters for the custom Pythia-160M SAEs.}
\end{table}
\end{document}